\documentclass[final]{cvpr}

\usepackage{times}
\usepackage{epsfig}
\usepackage{graphicx}
\usepackage{amsmath}
\usepackage{amssymb}
\usepackage{bm}
\usepackage{float}
\usepackage{multirow}
\usepackage{comment}
\usepackage{tabularx}
\usepackage{setspace}
\usepackage{booktabs}
\usepackage{setspace}
\usepackage{times}
\usepackage{epsfig}
\usepackage{graphicx}
\usepackage{amsmath}
\usepackage{amssymb}
\usepackage{subfigure} 
\usepackage{bbm}
\usepackage{float}
\usepackage{multirow}
\usepackage{tabularx}
\usepackage{setspace}
\usepackage{booktabs}
\usepackage{setspace}
\usepackage[pagebackref=false,breaklinks=true,colorlinks,bookmarks=false]{hyperref}

\def\cvprPaperID{6629} % *** Enter the CVPR Paper ID here
\def\confYear{CVPR 2021}

\begin{document}

%%%%%%%%% TITLE
\title{Dense Relation Distillation with Context-aware Aggregation for \\Few-Shot Object Detection}

\begin{comment}

\author{Hanzhe Hu\\
Institution1\\
Institution1 address\\
{\tt\small firstauthor@i1.org}
\and
Second Author\\
Institution2\\
First line of institution2 address\\
{\tt\small secondauthor@i2.org}
}
\end{comment}
\author{Hanzhe Hu\textsuperscript{1}, Shuai Bai\textsuperscript{2}, Aoxue Li\textsuperscript{1}, Jinshi Cui\textsuperscript{1}*, Liwei Wang\textsuperscript{1}* \\

\textsuperscript{1}Key Laboratory of Machine Perception (MOE), School of EECS, Peking University\\
\textsuperscript{2}Beijing University of Posts and Telecommunications\\
{\tt\small \{huhz, lax\}@pku.edu.cn  baishuai@bupt.edu.cn} 
{\tt\small \{cjs, wanglw\}@cis.pku.edu.cn}}
\maketitle
\pagestyle{empty}
\thispagestyle{empty}

%%%%%%%%% ABSTRACT
\begin{abstract}
Conventional deep learning based methods for object detection require a large amount of bounding box annotations for training, which is expensive to obtain such high quality annotated data. Few-shot object detection, which learns to adapt to novel classes with only a few annotated examples, is very challenging since the fine-grained feature of novel object can be easily overlooked with only a few data available. In this work, aiming to fully exploit features of annotated novel object and capture fine-grained features of query object, we propose Dense Relation Distillation with Context-aware Aggregation (DCNet) to tackle the few-shot detection problem. Built on the meta-learning based framework, Dense Relation Distillation module targets at fully exploiting support features, where support features and query feature are densely matched, covering all spatial locations in a feed-forward fashion. The abundant usage of the guidance information endows model the capability to handle common challenges such as appearance changes and occlusions. Moreover, to better capture scale-aware features, Context-aware Aggregation module adaptively harnesses features from different scales for a more comprehensive feature representation. Extensive experiments illustrate that our proposed approach achieves state-of-the-art results on PASCAL VOC and MS COCO datasets. Code will be made available at \href{https://github.com/hzhupku/DCNet}{https://github.com/hzhupku/DCNet}.
\end{abstract}

%%%%%%%%% BODY TEXT
\section{Introduction}
\renewcommand{\thefootnote}{\fnsymbol{footnote}}
\footnotetext[0]{*\ Corresponding authors.}
With the success of deep convolutional neural works, object detection has made great progress these years \cite{redmon2016you,ren2015faster,he2017mask}. The success of deep CNNs, however, heavily relies on large-scale datasets such as ImageNet \cite{deng2009imagenet} that enable the training of deep models. When the labeled data becomes scarce, CNNs can severely overfit and fail to generalize. While in contrast, human beings have exhibited strong performance in learning a new concept with only a few examples available. Since some object categories naturally have scarce examples or bounding box annotations are laborsome to obtain such as medical data. These problems have triggered increasing attentions to deal with learning models with limited examples. Few-shot learning aims to train models to generalize well with a few examples provided.  However, most existing few-shot learning works focus on image classification \cite{vinyals2016matching,snell2017prototypical,sung2018learning} problem and only a few focus on few-shot object detection problem. Since object detection not only requires class prediction, but also demands localization of the object, making it much more difficult than few-shot classification task.
\begin{figure}
    \centering
    \includegraphics[width=1\linewidth]{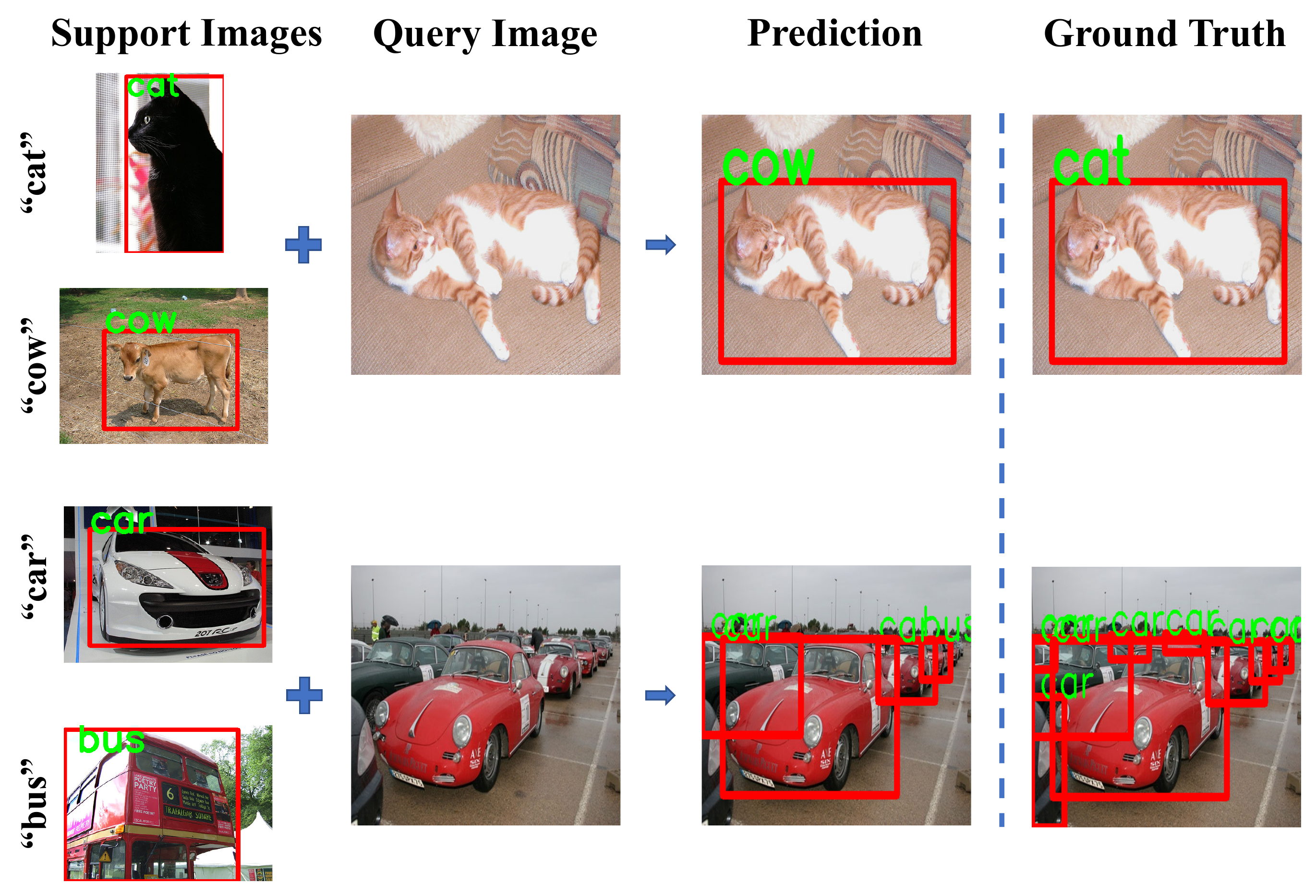}
    \caption{Two challenges for few-shot object detection. a) Appearance changes between support and query images are common, which results in a misleading manner. b) Occlusion problem brings about incomplete feature representation, causing false classification and missing detection. }
    \label{intro}
\end{figure}

Prior studies in few-shot object detection mainly consist of two groups. Most of them \cite{kang2019few,yan2019meta,xiao2020few} adopt a meta-learning \cite{finn2017model} based framework to perform feature reweighting for a class-specific prediction. While Wang \textit{et al.} \cite{wang2020frustratingly} adopt a two-stage fine-tuning approach with only fine-tuning the last layer of detectors and achieve state-of-the-art performance. Wu \textit{et al.} \cite{wu2020multi} also use similar strategy and focus on the scale variation problem in few-shot detection.

However, aforementioned methods often suffer from several drawbacks due to the challenging nature of few-shot object detection. Firstly, relations between support features and query feature are hardly fully explored in previous few-shot detection works, where global pooling operation on support features is mostly adopted to modulate the query branch, which is prone to loss of detailed local context. Specifically, appearance changes and occlusions are common for objects, as shown Fig.~\ref{intro}. Without enough discriminative information provided, the model is obstructed from learning critical features for class and bounding box predictions. Secondly, although scale variation problem has been widely studied in prior works \cite{lin2017feature,kim2018san,wu2020multi}, it remains a serious obstacle in few-shot detection tasks. Under few-shot settings, feature extractor with scale-aware modifications is inclined to overfitting, leading to a deteriorated performance for both base and novel classes.

%Specifically, Kang \textit{et al.} \cite{kang2019few} designs a few-shot detection model based on YOLO v2 \cite{redmon2017yolo9000} that learns generalizable meta features and automatically reweights the features for novel classes by producing class-specific activating coefficients from a few support samples. Yan \textit{et al.} \cite{yan2019meta} and Xiao \textit{et al.} \cite{xiao2020few} utilizes Faster R-CNN \cite{ren2015faster} detector and performs similar process as \cite{kang2019few}. While Wang \textit{et al.} \cite{wang2020frustratingly} simply adopts a two-stage fine-tuning approach and achieves state-of-the-art performance. And Wu \textit{et al.} \cite{wu2020multi}
%In this paper, we propose the dense relation distillation network with context-aware aggregation module (DCNet) to alleviate the above issues. Following similar settings from previous works, our proposed DCNet also adopts the meta-learning framework including the support branch and query branch. The overall framework of our proposed method is shown in Fig.\ref{framework}.
%Moreover, in order to endow model the ability to generalize to novel categories through exploiting abundant training data from base categories, we adopt the two-stage learning scheme which consists of base training and novel finetuning. 
In order to alleviate the above issues, we first propose the dense relation distillation module to fully exploit support set. Given a query image and a few support images from novel classes, the shared feature learner extracts query feature and support features for subsequent matching procedure. Intuitively, the criteria that determines whether query object and support object belong to the same category mainly measures how much feature similarity they share in common. When appearance changes or occlusions occur, local detailed features are dominant for matching candidate objects and template ones. Hence, instead of obtaining global representations of support set, we propose a dense relation distillation mechanism where query and support features are matched in a pixel-wise level. Specifically, key and value maps are produced from features, which serve as encoding visual semantics for matching and containing detailed appearance information for decoding respectively. With local information of support set effectively retrieved for guidance, the performance can be significantly boosted, especially in extremely low-shot scenarios. 
%The query feature and support features are then encoded into pairs of key and value maps. Specifically, key maps are used for measuring similarities between query feature and support features, which help determine where to retrieve relevant support values from. Hence key maps encode visual semantics for matching, while value maps contain detailed appearance information for the model to decode. Every pixel on the key feature maps of query and support images is densely matched over all the spatial locations and the produced matching scores are then used to address the value maps of support features. After combined with the value map of query feature, the output feature contains more semantic information exploited from support features, significantly benefiting the final class and bounding box predictions. When appearance changes or occlusions occur, local detailed features are indispensable for the prediction. While previous methods commonly harness support information by performing pooling operation to obtain an averaged form, leading to critical information loss. Our proposed dense matching fashion, on the other hand, is capable of fully exploiting support set in a pixel-level manner, thus effectively captures detailed information to benefit the learning process.

Furthermore, for the purpose of mitigating the scale variation problem, we design the context-aware feature aggregation module to capture essential cues for different scales during RoI pooling. Since directly modifying feature extractor could result in overfitting, we choose to perform adjustment from a more flexible perspective. Recognition of objects with different scales requires different levels of contextual information, while the fixed pooling resolution may bring about loss of substantial context information. Hence, an adaptive aggregation mechanism that allocates specific attention to local and global features simultaneously could help preserve contextual information for different scales of objects. Therefore, instead of performing RoI pooling with one fixed resolution, we choose three different pooling resolutions to capture richer context features. Then an attention mechanism is introduced to adaptively aggregate output features to present a more comprehensive representation. 

The contributions of this paper can be summarized as follows:
\begin{enumerate}
    \item We propose a dense relation distillation module for few-shot detection problem, which targets at fully exploiting support information to assist the detection process for objects from novel classes.
    \item We propose an adaptive context-aware feature aggregation module to better capture global and local features to alleviate scale variation problem, boosting the performance of few-shot detection. 
    \item Extensive experiments illustrate that our approach has achieved a consistent improvement on PASCAL VOC and MS COCO datasets. Specially, our approach achieves better performance than the state-of-the-art methods on the two datasets.
\end{enumerate}

%------------------------------------------------------------------------
\section{Related Work}
\subsection{General Object Detection}
Deep learning based object detection can be mainly divided into two categories: one-stage and two-stage detectors. One-stage detector YOLO series \cite{redmon2016you,redmon2017yolo9000,redmon2018yolov3} provide a proposal-free framework, which uses a single convolutional network to directly perform class and bounding box predictions. SSD \cite{liu2016ssd} uses default boxes to adjust to various object shapes. On the other hand, RCNN and its variants  \cite{girshick2014rich,he2015spatial,girshick2015fast,ren2015faster,he2017mask} fall into the second category. These methods first extract class-agnostic region proposals of the potential objects from a given image. The generated boxes are then further refined and classified into different categories by subsequent modules. Moreover, many works are proposed to handle scale variance \cite{lin2017feature,kim2018san,singh2018analysis,singh2018sniper}. Compared to one-stage methods, two-stage methods are slower but exhibit better performance. In our work, we adopt Faster R-CNN as the base detector. 
\subsection{Few-Shot Learning}
Few-shot learning aims to learn transferable knowledge that can be generalized to new classes with scarce examples. Bayesian inference is utilized in \cite{fei2006one} to generalize knowledge from a pretrained model to perform one-shot learning. %A hierarchical bayesian one-shot learning system is proposed in \cite{lake2013one}. Luo \textit{et al.} \cite{luo2017label} consider the problem of adapting to novel classes in a new domain. 
Meta-learning based methods have been prevalent in few-shot learning these days. Metric learning based methods \cite{koch2015siamese,vinyals2016matching,snell2017prototypical,sung2018learning} have achieved state-of-the-art performance in few-shot classification tasks.
%and they have the trait of being fast and predicting in a feed-forward manner. 
Matching Network \cite{vinyals2016matching} encodes input into deep neural features and performs weighted nearest neighbor matching to classify query images. Our proposed method is also based on matching mechanism. Prototypical Network \cite{snell2017prototypical} represents each class with one prototype which is a feature vector. Relation Network \cite{sung2018learning} learns a distance metric to compare the target image with a few labeled images. While optimization based methods \cite{ravi2016optimization,finn2017model} are proposed for fast adaptation to new few-shot task. \cite{HouCMSC19} proposes a cross-attention mechanism to learn correlations between support and query images. %Moreover, a number of works \cite{li2019revisiting,lifchitz2019dense} exploit local descriptors to reap additional knowledge from limited data. 
%In \cite{kim2019edge,gidaris2019generating}, graph neural network (GNN) is utilized to model relations between different categories.  
Above methods are focusing on the few-shot classification task while few-shot object detection problem is relatively under-explored.
\subsection{Few-Shot Object Detection}
Few-shot object detection aims to detect object from novel classes with only a few annotated training examples provided. LSTD \cite{chen2018lstd} and RepMet \cite{karlinsky2019repmet} adopt a general transfer learning framework which reduces overfitting by adapting pre-trained detectors to few-shot scenarios. %with limited training examples. 
%Following this framework, RepMet \cite{karlinsky2019repmet} adopts a distance metric learning classifier into the RoI classification head in the detector. 
Recently, Meta YOLO \cite{kang2019few} designs a novel few-shot detection model with YOLO v2 \cite{redmon2017yolo9000} that learns generalizable meta features and automatically reweights the features for novel classes by producing class-specific activating coefficients from support examples.
Meta R-CNN \cite{yan2019meta} and FsDetView \cite{xiao2020few} perform similar process with base detector as Faster R-CNN. 
%and achieves few-shot detection and segmentation. 
%FsDetView \cite{xiao2020few} adopts similar strategy as Meta R-CNN and further proposes novel aggregation module to boost the performance. 
TFA \cite{wang2020frustratingly} simply performs two-stage finetuning approach by only finetuning the classifier on the second stage and achieves better performance. MPSR \cite{wu2020multi} proposes multi-scale positive sample refinement to handle scale variance problem. %Moreover, these are a few works \cite{hsieh2019one,fan2020few} that explore relations between query and support set. 
CoAE \cite{hsieh2019one} proposes non-local RPN and focuses on one-shot detection from the view of tracking by comparing itself with other tracking methods, while our method performs cross-attention on features extracted by the backbone in a more straightforward way and targets at few-shot detection task. FSOD \cite{fan2020few} proposes attention-RPN, multi-relation detector and contrastive training strategy %which exploit the similarity between few-shot support set and query set 
to detect novel object. In our work, we adopt the similar meta-learning based framework as Meta R-CNN and further improve the performance. Moreover, with our proposed method, the class-specific prediction procedure can be successfully removed, simplifying the overall process. 
%------------------------------------------------------------------------
\begin{figure*}[!t]
    \centering
    \includegraphics[width=1\linewidth]{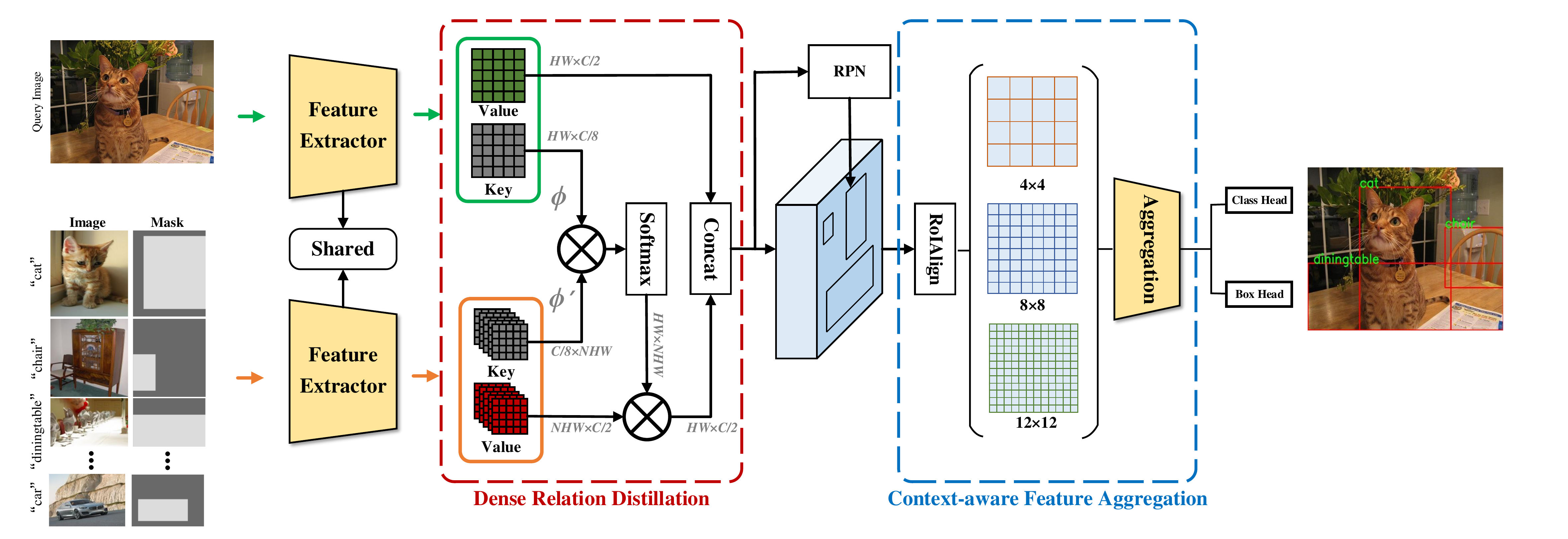}
    \caption{The overall framework of our proposed DCNet. For training, the input for each episode consists of a query image and N support image-mask pairs from N classes. The shared feature extractor first produces query feature and support features. Then, the dense relation distillation (DRD) module performs dense feature match to activate co-exisiting features of input query. With proposals produced by RPN, context-aware feature aggregation (CFA) module adaptively harnesses features generated with different scales of pooling operations, capturing different levels of features for a more comprehensive representation.  }
    \label{framework}
\end{figure*}
\section{Method}
%In this section, we will describe the proposed dense relation distillation network with context-aware feature aggregation (DCNet) in detail. Firstly, we will briefly introduce the problem definition of few-shot object detection, background of object detection and revisit a common solution to few-shot detection problem, which is also our baseline method. Then we will present the general framework of our proposed DCNet and explicitly introduce the proposed dense relation distillation network and context-aware aggregation module. Finally, we will bring out the learning scheme and supervision manner of the proposed model.
\subsection{Preliminaries.}
\noindent\textbf{Problem Definition.}
Following setting in \cite{kang2019few,yan2019meta}, object classes are divided into base classes $C_{\mathit{base}}$ with abundant annotated data and novel classes $C_{\mathit{novel}}$ with only a few annotated samples, where $C_{\mathit{base}}$ and $C_{\mathit{novel}}$ have no intersection. We aim to obtain a few-shot detection model with the ability to detect objects from both base and novel classes in testing by leveraging generalizable knowledge from base classes. The number of instances per category for novel classes is set as $k$ (\textit{i.e.}, $k$-shot). 

We align the training scheme with the episodic paradigm \cite{vinyals2016matching} for few-shot scenario. Given a $k$-shot learning task, each episode is constructed by sampling: 1) a support set containing image-mask pairs for different classes $S=\{x_i,y_i\}_{i=1}^N$, where $x_i \in \mathbb{R}^{h\times w\times 3}$ is an RGB image, $y_i \in \mathbb{R}^{h\times w}$ is a binary mask for objects of class $i$ in the support image generated from bounding box annotations and $N$ is the number of classes in the training set; 2) a query image $q$ and annotations $m$ for the training classes in the query image. The input to the model is the support pairs and query image, the output is detection prediction for query image.

\noindent\textbf{Basic Object Detection.}
%With the significant development in general object detection, 
The choice of base detectors is varied. \cite{kang2019few} utlizes YOLO v2 \cite{redmon2017yolo9000} which is a one-stage detector, while \cite{yan2019meta} adopts Faster R-CNN \cite{ren2015faster} which is a two-stage detector and provides consistently better results. Therefore, we also adopt Faster R-CNN as our base detector which consists of a feature extractor, region proposal network (RPN) and the detection head (RoI head).
\begin{comment}
Specifically, Faster R-CNN consists of the Region Proposal Network (RPN) and the detection head (RoI head). For a given image, the RPN head generates proposals with objectness scores and bounding box regression offsets. The loss function used in RPN is:
\begin{equation}
    L_{RPN} = \frac{1}{N_{obj}}\sum_i L_{bcls}^i + \frac{1}{N_{obj}}\sum_i L_{preg}^i.
\end{equation}
For the i-th anchor in a mini-batch, $L_{bcls}^i$ denotes the binary cross-entropy loss over foreground and background and $L_{preg}^i$ is the smooth $L_1$ loss for postive anchors defined in \cite{ren2015faster}. $N_{obj}$ is the total number of chosen anchors. These proposals are then fed into RoI head for feature extraction and detection predictions including class-specific scores and bounding-box offsets. The loss function used in RoI head is:
\begin{equation}
    L_{RoI} = \frac{1}{N_{RoI}}\sum_i L_{cls}^i + \frac{1}{N_{RoI}}\sum_i L_{preg}^i,
\end{equation}
where $L_{cls}$ is the normal cross-entropy class and $N_{RoI}$ is the number of RoIs in a mini-batch.
\end{comment}

\noindent\textbf{Feature Reweighting for Detection.}
We choose Meta-RCNN \cite{yan2019meta} as our baseline method. Formally, let $I$ denote an input query image, $\{I_{si},M_{si}\}|_{i=1}^N$ denote support images and masks converted from bounding-box annotations, where $N$ is the number of training classes. RoI features $z^j|_{j=1}^n$ is generated by the RoI pooling layer ($n$ is the number of RoIs) and class-specific vectors $w_i\in \mathbb{R}^C, i=1,2,...,N$ are produced with a reweighting module which shares its backbone parameters with the feature extractor, where $C$ is the feature dimension. Then class-specific feature $z_i$ is achieved with:
\begin{equation}
    z_i = z \otimes w_i, i=1,2,....,N,
\end{equation}
where $\otimes$ denotes channel-wise multiplication. Then class-specific prediction is performed to output the detection results. Based on this methodology, we further make a significant improvement and simplify the prediction procedure by removing the class-specific prediction.
\subsection{DCNet}
As illustrated in Fig.~\ref{framework}, we present the Dense Relation Distillation (DRD) module with Context-aware Feature Aggregation (CFA) module to fully exploit support features and capture essential context information. The two proposed components form the final model DCNet. We will first depict the architecture of the proposed DRD module. Then we will bring out the details of the CFA module.
%The input consists of a query image and support set, which are then fed into a shared feature extractor to produce the query feature and support features respectively. The two kinds of features are subsequently fed into the proposed dense relation distillation network to distill the relations between query and support samples. Distilled relation information is harnessed as a guidance for query feature to retrieve useful context from support features, which can be considered as a cross attention mechanism. With relation distillation, the refined query feature contains more concentrating representations of the target classes. After relation distillation, the attentive query feature is processed by RPN and RoI Align layer. During RoI Align process, it is much likely to lose important discriminative features due to the pooling operation. Hence the context-aware feature aggregation module is proposed to resolve this issue. Specifically, instead of performing pooling operation with a single size (8 in the original implementation), we empirically choose three pooling sizes for a more robust feature representation, which are 4, 8 and 12. Moreover, in order to possibly maximize the preservation of indispensable information, we introduce a attention mechanism to effectively fuse the three output features. Finally, the output is processed by the prediction module including a class head and a bounding-box regression head. The loss function we adopt is the same as the original implementation in Faster R-CNN \cite{ren2015faster}.
\subsubsection{Dense Relation Distillation Module }

\noindent\textbf{Key and Value Embedding. }Given a query image and support set, query and support features are produced by feeding them into the shared feature extractor. 
The input of the dense relation distillation (DRD) module is the query feature and support features. Both parts are first encoded into pairs of key and value maps through the dedicated deep encoders. The query encoder and support encoder adopt the same structure while not sharing parameters.

The encoder takes one or multiple feature as input and outputs two feature maps for each input feature: key and value with two parallel $3\times3$ convolution layers, which serve as reducing the dimension of the input feature to save computation cost.  Specifically, key maps are used for measuring the similarities between query feature and support features, which help determine where to retrieve relevant support values. Therefore, key maps are learned to encode visual semantics for matching and value maps store detailed information for recognition. Hence, for query feature, the output is a pair of key and value maps: $k_q \in \mathbb{R}^{C/8\times H \times W}, v_q \in \mathbb{R}^{C/2\times H \times W}$, where $C$ is the feature dimension, $H$ is the height, and $W$ is the width of input feature map. For support features, each of the features is independently encoded into key and value maps, the output is $k_s \in \mathbb{R}^{N\times C/8 \times H\times W}, v_s \in \mathbb{R}^{N\times C/2\times H \times W}$, where $N$ is the number of target classes (also the number of support samples). The generated key and value maps are further fed into the relation distillation part where keys maps of query and support are densely matched for addressing target objects.

%Specifically, key maps are used for measuring the similarities between query feature and support features, which help determine where to retrieve relevant support values. Therefore, key maps are learned to encode visual semantics for matching and value maps store detailed information for recognition. The generated key and value maps are further fed into the relation distillation part where keys maps of query and support are densely matched for addressing target objects.

\noindent\textbf{Relation Distillation. }After acquiring the key/value maps of query and support features, relation distillation is performed. As illustrated in Fig.~\ref{framework}, soft weights for value maps of support features are computed via measuring the similarities between key maps of query feature and support features. The pixel-wise similarity is performed in a non-local manner, formulated as:
\begin{equation}
    F(\bm{k_{qi}},\bm{k_{sj}}) = \phi(\bm{k_{qi}})^T \phi'(\bm{k_{sj}}),
\end{equation}
where $i$ and $j$ are the index of the query and support location, $\phi, \phi'$ denote two different linear transformations with parameters learned via back propagation during training process, forming a dynamically learned similarity function. After computing the similarity of pixel features, we perform softmax normalization to output the final weight $W$ :
\begin{equation}
    W_{ij} = \frac{exp(F(\bm{k_{qi}},\bm{k_{sj}}))}{\sum_j exp(F(\bm{k_{qi}},\bm{k_{sj}}))}.
\end{equation}
Then the value of the support features are retrieved by a weighted summation with the soft weights produced and then it is concatenated with the value map of query feature. Hence, the final output is formulated as:
\begin{equation}
    y = concat[v_q, W*v_s],
\end{equation}
where $*$ denotes matrix inner-product. Noted that there are $N$ support features, which brings $N$ key-value pairs. We perform summation over $N$ output results to obtain the final result, which is a refined query feature, activated by support features where there are co-existing classes of objects in query and support images.

Previous trials \cite{kang2019few,yan2019meta,xiao2020few} utilize class-wise vectors generated by global pooling of support features to modulate the query feature, which guide the feature learning from a holistic view. However, since appearance changes or occlusions are common in natural images, the holistic feature may be misleading when objects of the same class vary much between query and support samples. Also, when most parts of the objects are unseen due to the occlusions, the retrieval of local detailed features becomes substantial, which former methods completely neglect. Hence, equipped with the dense relation distillation module, pixel-level relevant information can be distilled from support features. As long as there exist some common characteristics, the pixels of query features belonging to the co-existing objects between query and support samples will be further activated, providing a robust modulated feature to facilitate the prediction of class and bounding-box. 

Our distillation method can be seen as an extension of the non-local self-attention mechanism \cite{vaswani2017attention,wang2018non}. However, instead of performing self-attention, we specially design the relation distillation model to realize information retrieval from support features to modulate the query feature, which can be treated as a cross attention.
\subsubsection{Context-aware Feature Aggregation}
\begin{figure}
    \centering
    \includegraphics[width=1\linewidth]{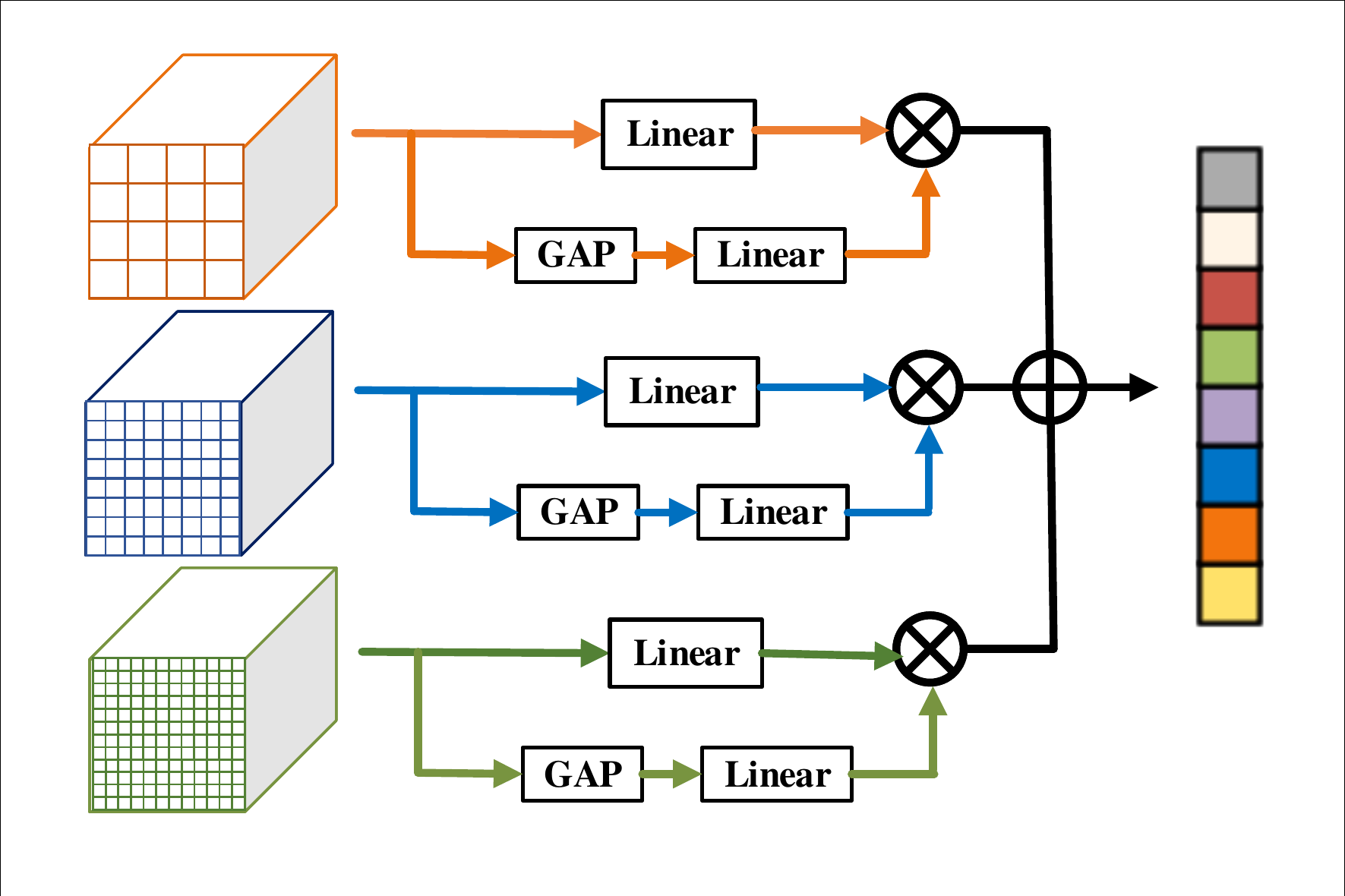}
    \caption{Illustration of context-aware feature aggregation. Attention mechanism is adopted to adaptively aggregate different features, where the weights are normalized with softmax function. }
    \label{cfa}
\end{figure}
After performing dense relation distillation, DRD module has fulfilled its duty. The refined query feature is subsequently fed into RPN where region proposals are output. Taking proposals and feature as input, RoI Align module performs feature extraction for final class prediction and bounding-box regression. Normally, pooling operation is implemented with a fixed resolution 8 in our original implementation, which is likely to cause information loss during training. For general object detection, this kind of information loss can be remedied with large scale of training data, while the problem becomes severe in few-shot detection scenarios with only a few training data available, which is inclined to induce a misleading detection results. Moreover, with scale variation amplified due to the few-shot nature, the model tends to lose the generalization ability to novel classes with adequate adaption to different scales. To this end, we propose Context-aware Feature Aggregation (CFA) module. Instead of using a fixed resolution 8, we empirically choose 4, 8 and 12 three resolutions and perform parallel pooling operation to obtain a more comprehensive feature representation. The larger resolution tends to focus on local detailed context information specially for smaller objects, while the smaller resolution targets at capturing holistic information to benefit the recognition of larger objects, providing a simple and flexible way to alleviate the scale variation problem.

Since each generated feature contains different level of semantic information. With the intention to efficiently aggregate features generated from different scales of RoI pooling, we further propose an attention mechanism to adaptively fuse the pooling results. As illustrated in Fig.~\ref{cfa}, we add an attention branch for each feature which consists of two blocks. The first block contains a global average pooling. The second one contains two consecutive fc layers. Afterwards, we add a softmax normalization to the generated weights for balancing the contribution of each feature. Then the final output of the aggregated feature is the weighted summation of the three features.

\begin{figure}
    \centering
    \includegraphics[width=1\linewidth]{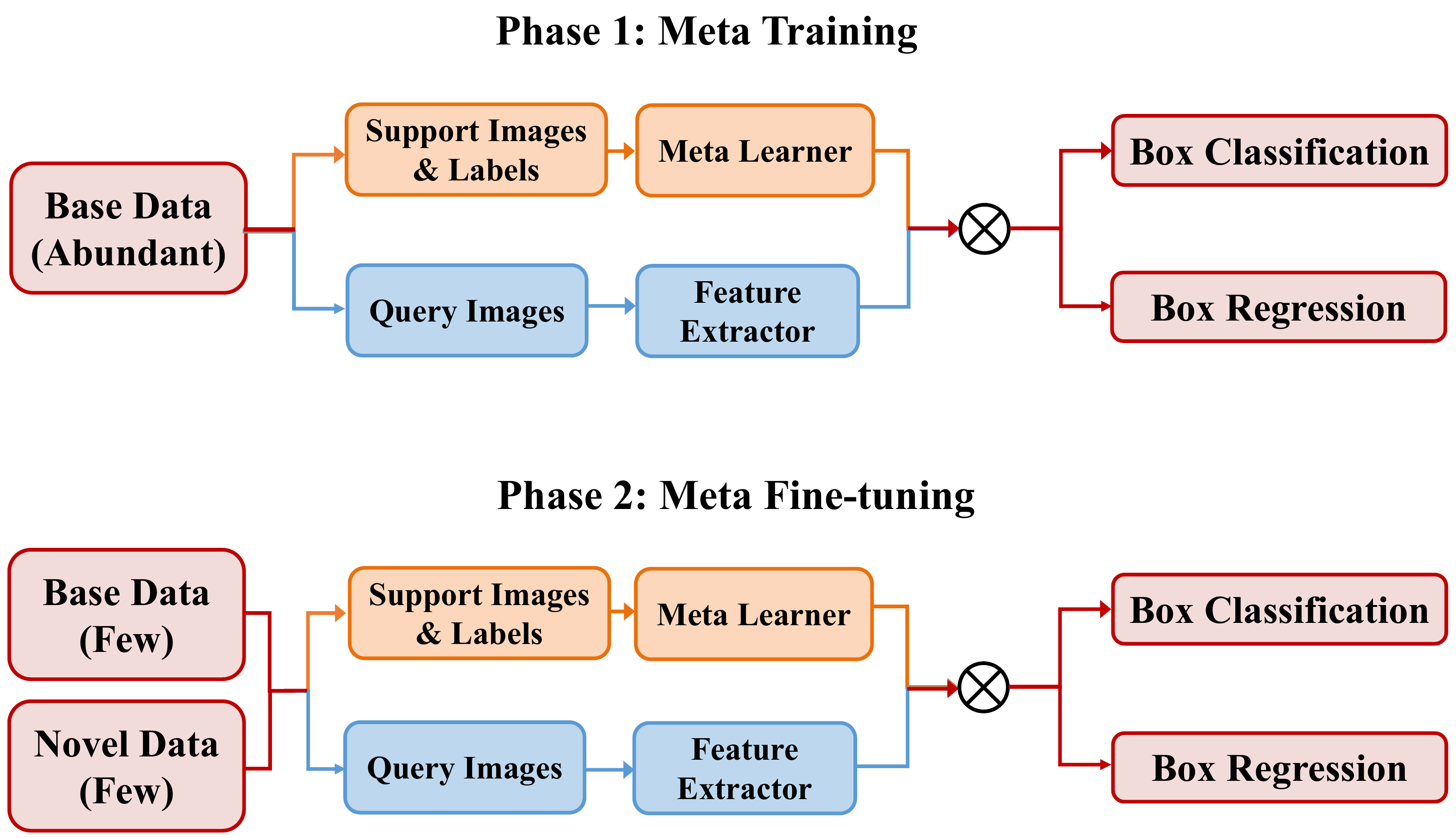}
    \caption{Demonstration of learning strategy of meta-learning based few-shot detection framework. The meta learner aims to acquire meta information and help the model to generalize to novel classes.} %A two-stage training paradigm (meta training and meta fine-tuning) is adopted with episodic learning.}
    \label{scheme}
\end{figure}

\subsection{Learning Strategy}
As illustrated in Fig.~\ref{scheme}, we follow the training paradigm in \cite{kang2019few,yan2019meta,xiao2020few}, which consists of meta-training and meta fine-tuning. In the phase of meta-training, abundant annotated data from base classes is provided. We jointly train the feature extractor, dense relation distillation module, context-aware feature aggregation module and other basic components of detection model. In meta fine-tuning phase, we train the model on both base and novel classes. As only $k$ labeled bounding-boxes are available for the novel classes, to balance between samples from base and novel classes, we also include $k$ boxes for each base class. The training procedure is the same as the meta-training phase but with fewer iterations for model to converge.
%------------------------------------------------------------------------
\section{Experiments}
In this section, %we evaluate our method on few-shot object detection by conducting a set of experiments on two datasets \textit{i.e.} PASCAL VOC and COCO. W
we first introduce the implementation details and experimental configurations in Sec.~\ref{setup}. Then we present our detailed experimental analysis on PASCAL VOC dataset in Sec.~\ref{voc} together with ablation studies and qualitative results. Finally, results on COCO dataset will be presented in Sec.~\ref{coco}. 
\subsection{Datasets and Settings}\label{setup}
Following the instructions in \cite{kang2019few}, we construct the few-shot detection datasets for fair comparison with other state-of-the-art methods. Moreover, to achieve a more stable few-shot detection results, we perform 10 random runs with different randomly sampled shots. Hence, all the results in the experiments is averaged results by 10 random runs.

\noindent\textbf{PASCAL VOC. } For PASCAL VOC dataset, we train our model on the VOC 2007 trainval and VOC 2012 trainval sets and test the model on VOC 2007 test set. The evaluation metric is the mean Average Precision (mAP). Both the trainval sets are split by object categories, where 5 are randomly chosen as novel classes and the left 15 are base classes. We use the same split as \cite{kang2019few}, where novel classes for four splits are \{``bird", ``bus", ``cow", ``motorbike" (``mbike"), ``sofa"\}, \{``aeroplane"(``aero", ``bottle", ``cow", ``horse", ``sofa"\}, \{``boat", ``cat", ``motorbike", ``sheep", ``sofa"\}, respectively. For few-shot object detection experiments, the few-shot dataset consists of images where $k$ object instances are available for each category and $k$ is set as 1/3/5/10.

\noindent\textbf{COCO.} MS COCO dataset has 80 object categories, where the 20 categories overlapped with PASCAL VOC are set to be novel classes. 5000 images from the validation set noted as minival are used for evaluation while the left images in the train and validation set are used for training. The process of constructing few-shot dataset is similar to PASCAL VOC dataset and $k$ is set as 10/30.

\noindent\textbf{Implementation Details. } We perform training and testing process on images with a single scale. The shorter side of the query image is resized to 800 pixels and longer sides are less than 1333 pixels while maintaining the aspect ratio. The support image is resized to a squared image of $256\times 256$. We adopt ResNet-101 \cite{he2016deep} as feature extractor and RoI Align \cite{he2017mask} as RoI feature extractor. The weights of the backbone is pre-trained on ImageNet \cite{deng2009imagenet}. After training on base classes, only the last fully-connected layer (for classification) is removed and replaced by a new one randomly initialized. It is worth noting that all parts of the model participate in learning process in the second meta fine-tuning phase without any freeze operation. We train our model with a mini-batch size as 4 with 2 GPUs. We utilize the SGD optimizer with the momentum of 0.9, and weight decay of 0.0001. For meta-training on PASCAL VOC, models are trained for 240k, 8k, and 4k iterations with learning rates of 0.005, 0.0005 and 0.00005 respectively. For meta fine-tuning on PASCAL VOC, models are trained for 1300, 400 and 300 iterations with learning rates as 0.005, 0.0005 and 0.00005 respectively. As for MS COCO dataset, during meta-training, models are trained for 56k, 14k and 10k iterations with learning rates of 0.005, 0.0005 and 0.00005 respectively. And during meta fine-tuning, model are trained for 2800, 700 and 500 iteration for 10-shot fine-tuning and 5600, 1400 and 1000 iterations for 30-shot fine-tuning.

\begin{table*}\small
\renewcommand\arraystretch{1.0}
\begin{center}
\begin{tabularx}{17.5cm} {p{2.5cm}|X<{\centering} X<{\centering} X<{\centering} X<{\centering} X<{\centering}|X<{\centering} X<{\centering} X<{\centering} X<{\centering} X<{\centering}|X<{\centering} X<{\centering} X<{\centering} X<{\centering} X<{\centering}}
\toprule[1.5pt]
   & \multicolumn{5}{c|}{Novel Set 1} & \multicolumn{5}{c|}{Novel Set 2} & \multicolumn{5}{c}{Novel Set 3}\\
  \midrule[1pt]
Methods / Shots  &  1 & 2 & 3 & 5 & 10 & 1 & 2 & 3 & 5 & 10 & 1 & 2 & 3 & 5 & 10 \\
\midrule[1pt]
\midrule[1pt]

LSTD \cite{chen2018lstd}   & 8.2 & 1.0 & 12.4 & 29.1 & 38.5 & 11.4 & 3.8 & 5.0 & 15.7 & 31.0 & 12.6 & 8.5 & 15.0 & 27.3 & 36.3\\
Meta YOLO \cite{kang2019few}   & 14.8 & 15.5 & 26.7 & 33.9 & 47.2 & 15.7 & 15.2 & 22.7 & 30.1 & 40.5 & 21.3 & 25.6 & 28.4 & 42.8 & 45.9\\
MetaDet* \cite{wang2019meta}   & 18.9 & 20.6 & 30.2 & 36.8 & 49.6 & 21.8 & 23.1 & 27.8 & 31.7 & 43.0 & 20.6 & 23.9 & 29.4 & 43.9 & 44.1\\
Meta R-CNN* \cite{yan2019meta}  & 19.9 & 25.5 & 35.0 & 45.7 & 51.5 & 10.4 & 19.4 & 29.6 & 34.8 & 45.4 & 14.3 & 18.2 & 27.5 & 41.2 & 48.1\\
TFA* w/fc \cite{wang2020frustratingly}  & 22.9 & 34.5 & 40.4 & 46.7 & 52.0 & 16.9 & 26.4 & 30.5 & 34.6 & 39.7 & 15.7 & 27.2 & 34.7 & 40.8 & 44.6\\
TFA* w/cos \cite{wang2020frustratingly}  & 25.3 & 36.4 & 42.1 & 47.9 & 52.8 & 18.3 & 27.5 & 30.9 & 34.1 & 39.5 & 17.9 & 27.2 & 34.3 & 40.8 & 45.6\\
FsDetView* \cite{xiao2020few}   & 24.2 & 35.3 & 42.2 & 49.1 & 57.4 & 21.6 & 24.6 & \textbf{31.9} & \textbf{37.0} & 45.7 & 21.2 & 30.0 & 37.2 & \textbf{43.8} & 49.6\\
\textbf{DCNet}*(ours)  & \textbf{33.9} & \textbf{37.4} & \textbf{43.7} & \textbf{51.1} & \textbf{59.6} & \textbf{23.2} & \textbf{24.8} & 30.6 & 36.7 & \textbf{46.6} & \textbf{32.3} & \textbf{34.9} & \textbf{39.7} & 42.6 & \textbf{50.7}\\
\bottomrule[1.5pt]
\end{tabularx}
\end{center}
\caption{Few-shot object detection performance on VOC 2007 test set of PASCAL VOC dataset. We report the mAP with IoU threshold 0.5 (AP50) under three different splits for five novel classes. * denotes the results averaged over multiple random runs.}
\label{vocsota}
\end{table*}

\begin{table}\small
\renewcommand\arraystretch{1.0}
\begin{center}
\begin{tabularx}{8.5cm} {p{3cm}|X<{\centering} X<{\centering} X<{\centering} X<{\centering} X<{\centering}}
\toprule[1.5pt]
   & \multicolumn{5}{c}{Novel Set 1} \\
  \midrule[1pt]
Methods / Shots  &  1 & 2 & 3 & 5 & 10  \\
\midrule[1pt]
\midrule[1pt]
%FRCN-ft  & 12.2 & 20.0 & 31.2 & 39.3 & 41.6 & 13.6 & 15.5 & 22.8 & 31.7 & 37.2 & 10.3 & 13.9 & 21.1 & 33.9 & 37.0\\
%FRCN-ft-full & 15.2 & 24.4 & 35.6 & 43.2 & 46.1 & 16.9 & 19.7 & 27.2 & 34.5 & 40.3 & 13.9 & 16.2 & 23.8 & 37.2 & 41.0\\
Meta R-CNN$\dagger$ & 21.8 & 27.9 & 38.6 & 45.0 & 51.4  \\
DCNet w/o CFA & 31.1 & 35.9 & 42.6 & 48.2 & 57.2 \\
Meta R-CNN$\dagger$ w/CFA & 22.0 & 31.3 & 39.6 & 45.6 & 52.6 \\
Meta R-CNN$\dagger$ w/CFA* & 24.2 & 32.0 & 40.2 & 47.8 & 53.0 \\
DCNet w/CFA & 32.9 & 36.8 & 43.1 & 49.1 & 57.6 \\
\textbf{DCNet}  & \textbf{33.9} & \textbf{37.4} & \textbf{43.7} & \textbf{51.1} & \textbf{59.6} \\
\bottomrule[1.5pt]
\end{tabularx}
\end{center}
\caption{Ablation study to evaluate the effectiveness of different components in our proposed method. The mAP with IoU threhold 0.5 (AP50) is reported. * denotes CFA module with attention aggregation fashion. $\dagger$ denotes our implementation.}
\label{ablation}
\end{table}

\noindent\textbf{Baseline Method.} 
%We choose several baseline methods for comparison with our proposed method. Since we adopt Faster R-CNN (FRCN) as our base detector, and our model can be seen as a meta-learning extension to FRCN, we perform several baselines based on FRCN. Specifically, 
%\textbf{FRCN-joint} denotes that FRCN is jointly trained on data from base classes and novel classes with the same number of iterations as our model. \textbf{FRCN-ft} indicates that FRCN adopts a two-stage training strategy. In the base training phase, base classes training examples are used, while in the novel fine-tuning stage, the combination of base classes and novel classes is utilized. The number of iterations is the same as our model. \textbf{FRCN-ft-full} adopts the same training protocol as FRCN-ft while trains the model to fully converge in the second phase. Moreover, we also implement a meta-learning based method as our baseline, which is \textbf{Meta R-CNN} \cite{yan2019meta}.
Since we adopt Faster-RCNN as base detector, we choose Meta R-CNN \cite{yan2019meta} as the baseline method. Moreover, we implement it by ourselves for a more fair comparison.

\subsection{Experiments on PASCAL VOC}\label{voc}
In this section, we conduct experiments on PASCAL VOC dataset. We first compare our method with the state-of-the-art methods. Then we carry out ablation studies to perform comprehensive analysis of the components of our proposed DCNet. Finally, some qualitative results are presented to provide an intuitive view of the validity of our method. For all the experiments, we run 10 trials with random support data and report the averaged performance.

\subsubsection{Comparisons with State-of-the-art Methods}
In Table \ref{vocsota}, we compare our method with former state-of-the-art methods which mostly report results with multiple random runs. Our proposed DCNet achieves state-of-the-art results on almost all the splits with different shots and outperforms previous methods by a large margin. Specifically, in extremely low-shot settings (\textit{i.e.} 1-shot), our method outperforms others by about 10\% in split 1 and 3, providing a convincing proof that our DCNet is able to capture local detailed information to overcome the variations brought by the randomly sampled training shots.

\begin{table}\small
\renewcommand\arraystretch{1.0}
\begin{center}
\begin{tabularx}{8cm} {p{3cm}<{\centering}|X<{\centering} X<{\centering} X<{\centering} |p{1.5cm}<{\centering}}
\toprule[1.5pt]
Methods / Resolution  &  4 & 8 & 12 &  10-shot  \\
\midrule[1pt]
\midrule[1pt]
%FRCN-ft  & 12.2 & 20.0 & 31.2 & 39.3 & 41.6 & 13.6 & 15.5 & 22.8 & 31.7 & 37.2 & 10.3 & 13.9 & 21.1 & 33.9 & 37.0\\
%FRCN-ft-full & 15.2 & 24.4 & 35.6 & 43.2 & 46.1 & 16.9 & 19.7 & 27.2 & 34.5 & 40.3 & 13.9 & 16.2 & 23.8 & 37.2 & 41.0\\
DCNet    & \checkmark & - & - & 56.8  \\
DCNet   & - & \checkmark & - & 57.2  \\
DCNet   & - & - & \checkmark & 58.7  \\
DCNet   & \checkmark & \checkmark & - & 57.9 \\
DCNet   & - & \checkmark & \checkmark & 59.1  \\
DCNet   & \checkmark & - & \checkmark & 58.9  \\
DCNet   & \checkmark & \checkmark & \checkmark & \textbf{59.6} \\
\bottomrule[1.5pt]
\end{tabularx}
\end{center}
\caption{The impact of different RoI pooling resolutions. The experiments are conducted on VOC 2007 test set of PASCAL VOC dataset with novel split1 and AP50 on 10-shot task averaged from 10 random runs is reported.}
\label{resolution}
\end{table}
\subsubsection{Ablation Study}
We present results of comprehensive ablation studies to analyze the effectiveness of various components of the proposed DCNet. All ablation studies are conducted on the PASCAL VOC 2007 test set with the first novel splits. All results are averaged over 10 random runs.

%\noindent\textbf{Impact of meta-learning paradigm}
%We first compare finetune based methods FRCN-ft and FRCN-ft-full with meta-learning based method Meta R-CNN to demonstrate the advantage of meta-learning paradigm. Based on the first three lines of Table \ref{ablation}, Meta-RCNN significantly improves the performance compared with FRCN-ft and FRCN-ft-full, by simply deploying a reweighting module to modulate query feature to generalize to novel classes. Moreover, the introduction of meta-learner provides the detector an additional technique to obtain guidance of novel classes, thus boosting the generalization ability for base detector. 

\begin{table*}\small
\renewcommand\arraystretch{1.0}
\begin{center}
\begin{tabularx}{17.5cm} {p{0.6cm}|p{2.5cm}|X<{\centering} X<{\centering} X<{\centering} X<{\centering} X<{\centering}X<{\centering}| X<{\centering} X<{\centering} X<{\centering} X<{\centering} X<{\centering} X<{\centering}}
\toprule[1.5pt]
  & & \multicolumn{6}{c|}{Average Precision} & \multicolumn{6}{c}{Average Recall} \\
  \midrule[1pt]
\centering{Shots} & Methods  &  AP & AP$_{50}$ & AP$_{75}$ & AP$_S$ & AP$_M$ & AP$_L$ & AR$_1$ & AR$_{10}$ & AR$_{100}$ & AR$_S$ & AR$_M$ & AR$_L$  \\
\midrule[1pt]
\midrule[1pt]

\multirow{8}{*}{10} & LSTD \cite{chen2018lstd}   & 3.2 & 8.1 & 2.1 & 0.9 & 2.0 & 6.5 & 7.8 & 10.4 & 10.4 & 1.1 & 5.6 & 19.6 \\
~ & Meta YOLO \cite{kang2019few}   & 5.6 & 12.3 & 4.6 & 0.9 & 3.5 & 10.5 & 10.1 & 14.3 & 14.4 & 1.5 & 8.4 & 28.2 \\
~ & MetaDet* \cite{wang2019meta}   & 7.1 & 14.6 & 6.1 & 1.0 & 4.1 & 12.2 & 11.9 & 15.1 & 15.5 & 1.7 & 9.7 & 30.1 \\
~ & Meta R-CNN* \cite{yan2019meta}  & 8.7 & 19.1 & 6.6 & 2.3 & 7.7 & 14.0 & 12.6 & 17.8 & 17.9 & 7.8 & 15.6 & 27.2 \\
~ & TFA* w/fc \cite{wang2020frustratingly}  & 9.1 & 17.3 & 8.5 & - & - & - & - & - & - & - & - & -\\
~ & TFA* w/cos \cite{wang2020frustratingly}  & 9.1 & 17.1 & 8.8 & - & - & - & - & - & - & - & - & - \\
~ & FsDetView* \cite{xiao2020few}   & 12.5 & 27.3 & 9.8 & 2.5 & 13.8 & 19.9 & 20.0 & 25.5 & 25.7 & 7.5 & 27.6 & 38.9 \\
~ & \textbf{DCNet}*(ours)  & \textbf{12.8} & 23.4 & \textbf{11.2} & \textbf{4.3} & \textbf{13.8} & \textbf{21.0} & 18.1 & \textbf{26.7} & 25.6 & \textbf{7.9} & 24.5 & 36.7 \\
\midrule[1pt]
\multirow{8}{*}{30} & LSTD \cite{chen2018lstd}   & 6.7 & 15.8 & 5.1 & 0.4 & 2.9 & 12.3 & 10.9 & 14.3 & 14.3 & 0.9 & 7.1 & 27.0 \\
~ & Meta YOLO \cite{kang2019few}   & 9.1 & 19.0 & 7.6 & 0.8 & 4.9 & 16.8 & 13.2 & 17.7 & 17.8 & 1.5 & 10.4 & 33.5 \\
~ & MetaDet* \cite{wang2019meta}   & 11.3 & 21.7 & 8.1 & 1.1 & 6.2 & 17.3 & 14.5 & 18.9 & 19.2 & 1.8 & 11.1 & 34.4 \\
~ & Meta R-CNN* \cite{yan2019meta}  & 12.4 & 25.3 & 10.8 & 2.8 & 11.6 & 19.0 & 15.0 & 21.4 & 21.7 & 8.6 & 20.0 & 32.1 \\
~ & TFA* w/fc \cite{wang2020frustratingly}  & 12.0 & 22.2 & 11.8 & - & - & - & - & - & - & - & - & -\\
~ & TFA* w/cos \cite{wang2020frustratingly}  & 12.1 & 22.0 & 12.0 & - & - & - & - & - & - & - & - & - \\
~ & FsDetView* \cite{xiao2020few}   & 14.7 & 30.6 & 12.2 & 3.2 & 15.2 & 23.8 & 22.0 & 28.2 & 28.4 & 8.3 & 30.3 & 42.1 \\
~ & \textbf{DCNet}*(ours)  & \textbf{18.6} & \textbf{32.6} & \textbf{17.5} & \textbf{6.9} & \textbf{16.5} & \textbf{27.4} & \textbf{22.8} & 27.6 & \textbf{28.6} & \textbf{8.4} & 25.6 & \textbf{43.4} \\
\bottomrule[1.5pt]

\end{tabularx}
\end{center}
\caption{Few-shot object detection performance on COCO minival of MS COCO dataset. We report the mean Averaged Precision and mean Averaged Recall on the 20 novel classes of COCO. * denotes the results averaged over multiple random runs.}
\label{cocosota}
\end{table*}

\begin{figure*}
    \centering
    \includegraphics[width=0.9\linewidth]{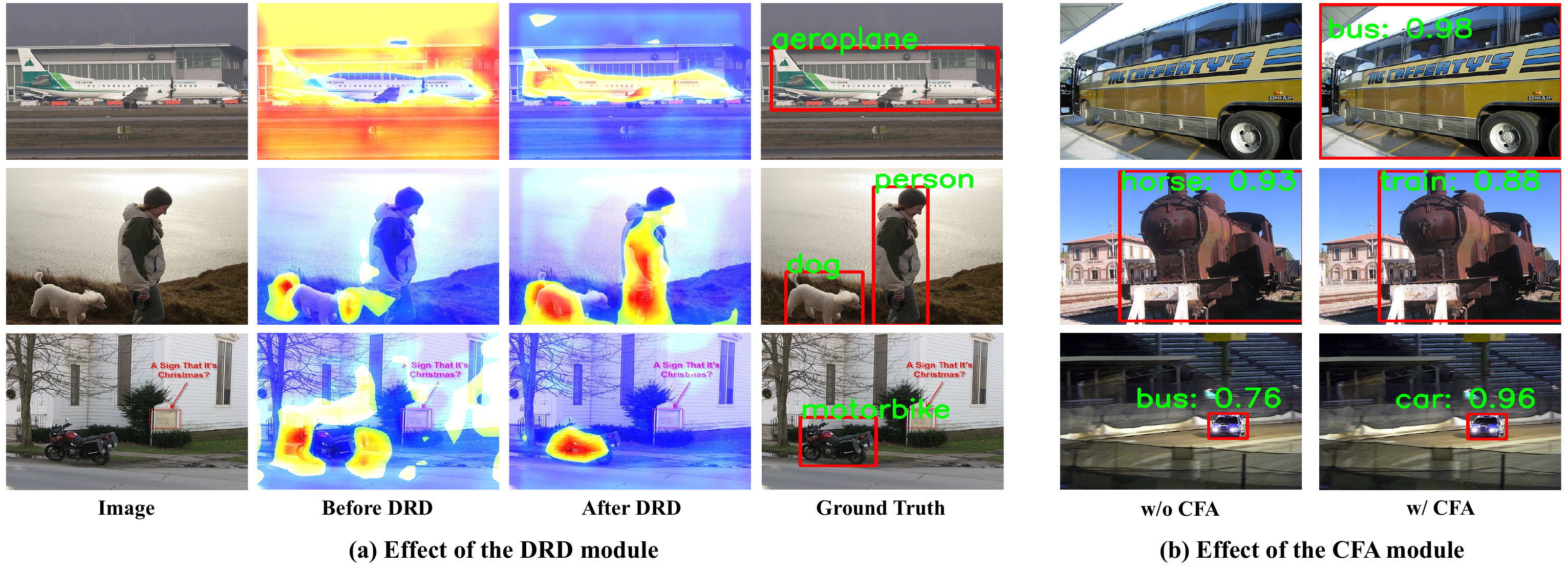}
    \caption{(a). Visualizations of features before and after dense relation distillation module. (b). Visualizations of effect of context-aware feature aggregation module.}
    \label{vis_det}
\end{figure*}

\noindent\textbf{Impact of dense relation distillation module. }
We conduct experiments to validate the superiority of the proposed dense relation distillation (DRD) module. Specifically, we implement the baseline method for meta-learning based few-shot detection Meta R-CNN with class-specific prediction for the final box classification and regression. While the DRD module requires no extra  class-specific processing. As shown in line 1 and 2 of Table \ref{ablation}, DCNet w/o CFA equals to Faster R-CNN equipped with DRD module,  our proposed DRD module achieves consistent improvement on all novel splits with all shots number, which effectively demonstrates the supremacy of the relation distillation mechanism over the baseline method. Moreover, the improvement over baseline is significant when the shot number is low, which proves that the DRD module successfully exploits useful information from limited support data.

\noindent\textbf{Impact of context-aware feature aggregation module. }
We carry out experiments to evaluate the validity of the proposed context-aware feature aggregation (CFA) module. Specifically, RoI features generated from parallel branches are aggregated with a simple summation. From line 1 and 3 of the table, with the introduction of CFA module, Meta R-CNN achieves notable gains over the baseline. Since CFA module targets at preserving detailed information in a scale-aware manner, different levels of detailed features can be retrieved to assist the prediction process.

\noindent\textbf{Impact of different RoI pooling resolutions. }
To further evaluate the impact of different RoI pooling resolutions, we perform explicit experiments to show the detailed performance. As shown in Table \ref{resolution}, solely adopting larger pooling resolution could yield better performance. However, only when aggregating features generated with all three resolutions, the best performance could be obtained.

\noindent\textbf{Impact of attentive aggregation fashion for CFA module. }
Based on the plain CFA module, we further propose an attention-based aggregation mechanism to adaptively fuse different RoI features. As presented in line 3 and line 4 of Table \ref{ablation}, the attention aggregation mechanism can further boost the performance of the model, which promotes the plain CFA module with a more comprehensive feature representation, effectively balancing the contributions of each extracted features. Finally, with the combination of DRD module and CFA module, we present DCNet, which achieves the best performance according to Table \ref{ablation}.
\vspace{-0.4cm}
\subsubsection{Qualitative Results}
%\vspace{-0.15cm}
To further comprehend the effect of dense relation distillation (DRD) module, we visualize features before and after DRD module. As shown in Fig.~\ref{vis_det} (a), after relation distillation, query features can be activated to facilitate the subsequent detection procedure. Moreover, different from former meta-learning based methods which performs prediction in a class-wise manner, our proposed DRD module can model relations between query and support features in all classes at the same time as shown in the second line of Fig.~\ref{vis_det} (a). The DRD module enables the model to focus more on the query objects under the guidance of support information. Additionally, we also visualize the effect of CFA module presented in Fig.~\ref{vis_det} (b). With a relatively large or small query object as input, DCNet w/o CFA suffers from false classification or missing detection , while the introduction of CFA module could effectively resolve this issue. 
%\vspace{-0.15cm}
\subsection{Experiments on MS COCO}\label{coco}

We evaluate 10/30-shot setups on MS COCO benchmark and report the averaged performance with the standard COCO metrics over 10 runs with random shots. The results on novel classes can be seen in Table \ref{cocosota}. Despite the challenging nature of COCO dataset with large number of categories, our proposed DCNet achieves state-of-the-art performance on most of the metrics.
%------------------------------------------------------------------------
%\vspace{-0.25cm}
\section{Conclusions}
\vspace{-0.15cm}
In this paper, we have presented the Dense Relation Distillation Network with Context-aware Aggregation (DCNet) to tackle few-shot object detection problem. Dense relation distillation module adopts dense matching strategy between query and support features to fully exploit support information. Furthermore, context-aware feature aggregation module adaptively harnesses features from different scales to produce a more comprehensive feature representation. The ablation experiments demonstrate the effectiveness of each component of DCNet. Our proposed DCNet achieves state-of-the-art results on two benchmark datasets, \textit{i.e.} PASCAL VOC and MS COCO. 

\section*{Acknowledgment}
This work was supported by the National Key R\&D Program of China under grant 2017YFB1002804.

{\small
\bibliographystyle{ieee_fullname}
\bibliography{egbib}
}

\end{document}